\documentclass[runningheads]{llncs}
\usepackage{marvosym}
\usepackage[T1]{fontenc}
\usepackage{graphicx}
\usepackage{esvect}
\usepackage{amsmath}
\usepackage{amssymb}
\usepackage{enumitem}
\usepackage{cite}
\usepackage[colorlinks=true, allcolors=blue]{hyperref}
\usepackage{cleveref}
\usepackage{booktabs}

\begin{document}
\title{Exemplar-Guided Planning: Enhanced LLM Agent for KGQA}
%
%
\author{Jingao Xu \and
Shuoyoucheng Ma \and
Xin Song \and \\
Rong Jiang \and
Hongkui Tu \and
Bin Zhou\textsuperscript{(\Letter)}}
\authorrunning{J. Xu et al.}
%
\institute{College of Computer Science and Technology, National University of Defense Technology, Changsha, China\\
\email{binzhou@nudt.edu.cn}\\}
\maketitle              
\begin{abstract}

Large Language Models (LLMs) as interactive agents show significant promise in Knowledge Graph Question Answering (KGQA) but often struggle with the semantic gap between natural language queries and structured knowledge graph (KG) representations. This leads to suboptimal planning and inefficient exploration on KG, while training-free approaches often underutilize valuable reasoning patterns in training data. To address these limitations, we propose a novel framework, Exemplar-Guided Planning (EGP), which enhances the planning capabilities of LLM agents for KGQA. EGP first preprocesses the training set questions via entity templating to normalize semantic variations. It then retrieves highly similar exemplary questions and their successful reasoning paths from this preprocessed set using semantic embeddings and an efficient FAISS index. These retrieved exemplars dynamically guide the LLM's planning process in two key phases: (1) Task Decomposition, by aligning generated sub-objectives with proven reasoning steps, and (2) Relation Exploration, by providing high-quality auxiliary information to improve relation pruning accuracy. Additionally, we introduce a Smart Lookahead mechanism during relation exploration to improve efficiency by preemptively exploring promising paths and potentially terminating exploration earlier. We apply EGP to the Plan-on-Graph (PoG) framework, termed PoG-EGP. Extensive experiments on two real-world KGQA datasets, WebQSP and CWQ, demonstrate that PoG-EGP significantly improves over the baseline PoG system and other compared methods.

\end{abstract}
\section{Introduction}
Large Language Models (LLMs) have recently demonstrated remarkable capabilities in numerous NLP tasks, driven by pre-training on massive text corpora. However, they face limitations such as hallucinations, static knowledge, and opacity, particularly in knowledge-intensive applications like Knowledge Graph Question Answering (KGQA). Meanwhile, Knowledge Graphs (KGs), with their explicit, structured, and verifiable knowledge, offer key complementary advantages, including easier updates and more transparent reasoning processes.

Integrating LLMs and KGs has spurred significant research in KGQA. Some approaches utilized KGs as external knowledge for Retrieval-Augmented Generation (RAG)\cite{xu2024retrieval, li2024simple} or employed LLMs to translate questions into formal KG queries (e.g., SPARQL)\cite{luo2023chatkbqa, shah2024improving}. A key limitation of these methods is that LLMs do not directly participate in the KG reasoning process. This hinders their ability to perform complex, multi-hop logical inference over the KG's structure, especially for intricate problems\cite{sun2023think}.

A new paradigm has emerged to overcome these challenges: LLMs as agents that interactively explore and reason over KGs\cite{sun2023think}. These agents can navigate the KG, retrieve information, evaluate intermediate results, and adapt their reasoning strategies dynamically. Plan-on-Graph (PoG)\cite{chen2024pog} is a notable example, introducing memory and reflection for adaptive, self-correcting path exploration. Such agent-based approaches enable deeper engagement with the KG structure, improving reasoning depth, accuracy, and interpretability.

\begin{figure}[!htbp]
\centering
\includegraphics[width=\textwidth]{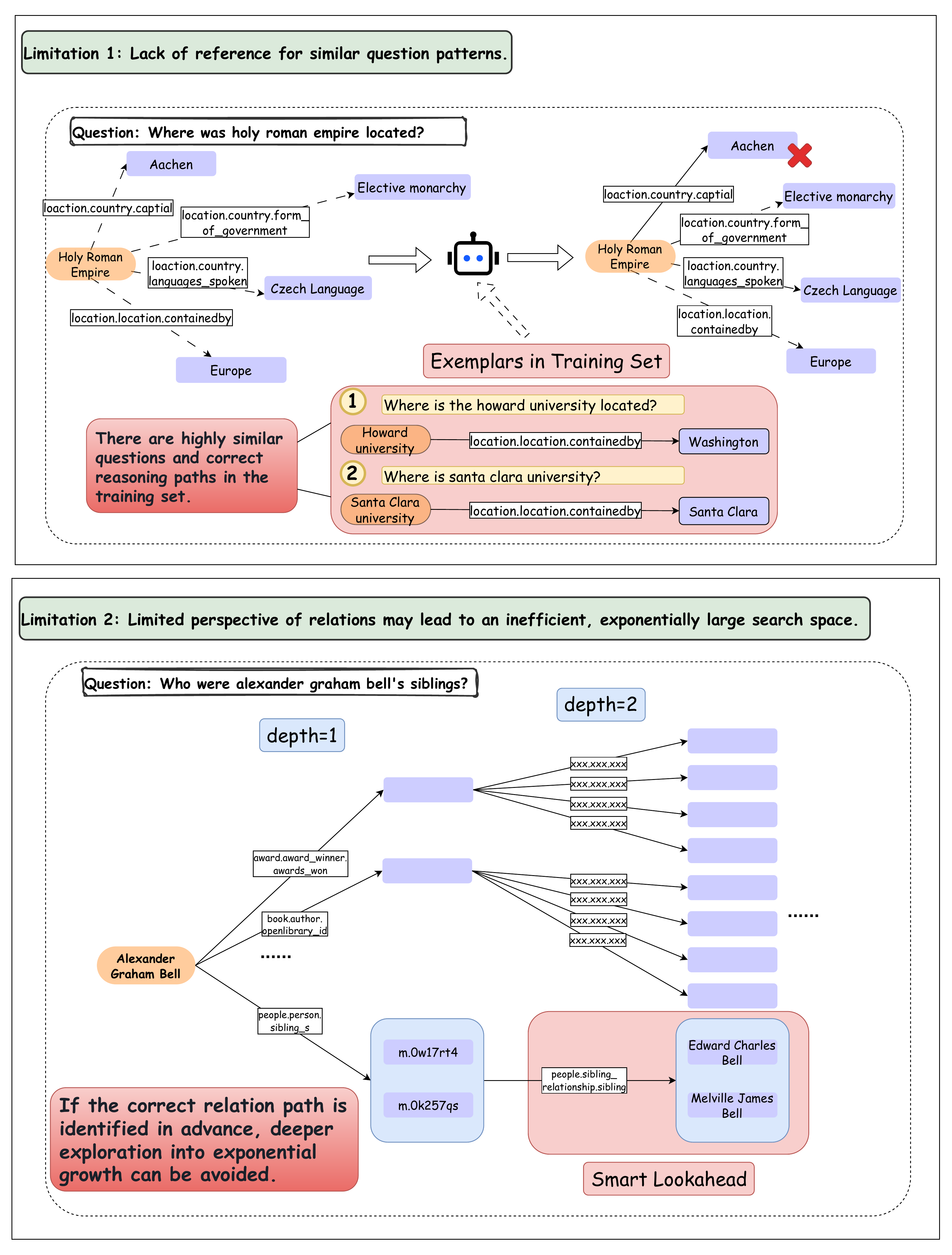}
\caption{Illustration of key limitations in existing training-free KGQA agents that motivate our Exemplar-Guided Planning (EGP) framework: (1) Underutilization of reasoning path patterns from similar questions available in training data, which can lead to incorrect answers. (2) Inefficient exploration due to a limited perspective of relations without advanced guidance mechanisms, potentially resulting in an unnecessarily large or suboptimal search space.} \label{Limitation}
\end{figure}

However, existing research has predominantly focused on leveraging LLMs’ inherent capabilities, with less direct attention given to fully bridging the semantic gap between LLMs and KG structural representations\cite{liu2025symagent}, leading to erroneous exploration for complex problems\cite{chen2025rsrule}. As shown in Fig. \ref{Limitation}, training-free methods like ToG\cite{sun2023think} and PoG\cite{chen2024pog}, while accessible, fail to leverage valuable reasoning path patterns in training sets that are enlightening and could bridge the LLM-KG semantic gap. Additionally, the exponential growth of exploration space with depth forces wider search beams, increasing the cost of answering the question.

We propose Exemplar-Guided Planning (EGP) to address these limitations and to enhance LLMs’ planning in interactive KG reasoning. EGP preprocesses training questions through entity templating, creates semantic embeddings, and retrieves similar exemplary questions to guide LLM planning. These exemplars, retrieved for their high similarity to the input question, are expected to employ reasoning paths that are highly similar or even identical. We implement EGP with PoG\cite{chen2024pog} as PoG-EGP. EGP guides LLMs in: (1) Task Decomposition: using exemplary reasoning paths to better understand KG hierarchies and align sub-objectives with necessary reasoning steps; and (2) Relation Exploration: providing high-quality auxiliary information to improve relation selection accuracy. We also introduce a Smart Lookahead mechanism to examine next-hop relations for paths similar to exemplars, avoiding deeper explorations. Extensive experiments on two KGQA datasets validate EGP’s effectiveness. The main contributions of this paper are as follows:

\begin{itemize}
    \item We propose the EGP framework that leverages exemplary questions and reasoning paths from training data to provide high-quality auxiliary information for LLMs, enhancing their planning capabilities in interactive KG reasoning by helping them understand logical and structural patterns within the KG.
    \item We introduce a Smart Lookahead mechanism that strategically limits exploration depth in certain cases, preventing exponential growth of the search space and improving computational efficiency.
    \item We demonstrate through extensive experiments on WebQSP and CWQ datasets that our PoG-EGP implementation significantly improves performance and efficiency compared to the baselines.
\end{itemize}

\section{Related Work}
\subsection{LLMs for KGQA}
LLMs, with their strong language understanding and reasoning, are increasingly vital in KGQA. Early approaches often used KGs as external knowledge: some retrieved KG facts to augment LLM context for answer generation\cite{xu2024retrieval, li2024simple}, while others used LLMs to convert natural language questions into formal queries like SPARQL\cite{luo2023chatkbqa, shah2024improving}. A common limitation was the LLM's passive role in the actual KG reasoning process.

A more recent paradigm treats LLMs as agents that interactively explore KGs. These agents actively navigate the KG, discover information, and perform multi-hop reasoning. ToG\cite{sun2023think} is a notable example, employing LLMs for iterative beam search over KG paths. Plan-on-Graph (PoG)\cite{chen2024pog}  advanced this by introducing guidance, memory, and reflection mechanisms, enabling adaptive exploration width and self-correction of erroneous paths, thus improving planning and reasoning. Our work builds upon PoG, applying the EGP framework to further enhance its planning capabilities. SymAgent\cite{liu2025symagent} also explores LLM agents on KGs, focusing on deriving symbolic rules from KGs to guide question decomposition, sharing a similar motivation with our EGP framework in leveraging structural patterns for better planning.

\subsection{Semantic Retrieval}
Semantic retrieval is crucial for identifying relevant information beyond keyword matching by understanding the deeper meaning and intent within text. Text embedding models are central to this, converting text into dense vector representations where semantic similarity corresponds to vector proximity. Transformer-based models like BERT\cite{devlin2019bert} and its variants are standard for generating high-quality embeddings due to their self-attention mechanisms capturing rich contextual information. Sentence-BERT\cite{reimers1908sentence} specifically fine-tunes these models for sentence similarity. The BGE model\cite{bge_embedding}, trained through a multi-stage process including contrastive learning, has shown strong performance. For efficient retrieval from large-scale embedding sets, FAISS\cite{douze2024faiss} is a widely adopted library. Our EGP framework leverages these advancements, using bge-large-en-v1.5 for text embedding and FAISS for efficient retrieval of similar exemplary questions.

\section{Preliminary}

\textbf{Knowledge Graph(KG)} is a semantic network used to represent structured knowledge, modeling concepts, entities, and their semantic relations in the real world in a graph format. A KG contains a large number of factual triples, with concepts or entities as nodes and relations or attributes as edges. It can be formally modeled as: $G = \{ (e, r, e') \mid e, e' \in E,\, r \in R \}$, where E and R denote the set of entities and relations, respectively.

\textbf{Relation Paths} are sequences of relations $z = \{ r_1, r_2, \ldots, r_l \}$, where each $r_i \in R$, and $l$ represents the length of the path. Such a relation path is considered realizable in the KG if there exists a corresponding sequence of entities $e_0, e_1, \ldots, e_l \in E$ such that for all $i \in [1, l]$, the triple $(e_{i-1}, r_i, e_i)$ is present in $G$.

\textbf{Reasoning Paths} are instantiations of a realizable relation path $z$ in the KG: $p_z = e_0 \xrightarrow{r_1} e_1 \xrightarrow{r_2} \cdots \xrightarrow{r_l} e_l$, where $e_i \in E$ denotes the i-th entity and $r_i$ denotes the i-th relation in the relation path $z$.

\textbf{Knowledge Graph Question Answering(KGQA)} is a reasoning task over a KG, which aims to answer natural language questions based on the factual information provided in the KG. Given a natural language question $q$, the topic entities $T_q$ mentioned in $q$, and a KG $G$, the goal of KGQA is to generate the answer(s) $A_q$ to question $q$. Following previous studies\cite{sun2024think, chen2024pog}, we assume any entity $e_q \in T_q$ mentioned in $q$ and answers $a_q \in A_q$ are labeled and linked to the corresponding entities in G, i.e., $T_q, A_q \subseteq E$.

\section{Method}

In this section, we introduce the technical details of our proposed novel exemplar-guided enhancement framework EGP, designed to improve the planning capabilities of LLMs in interactive KG reasoning methods.

\begin{figure}[tbp]
\centering
\includegraphics[width=\textwidth, height=0.43\textheight]{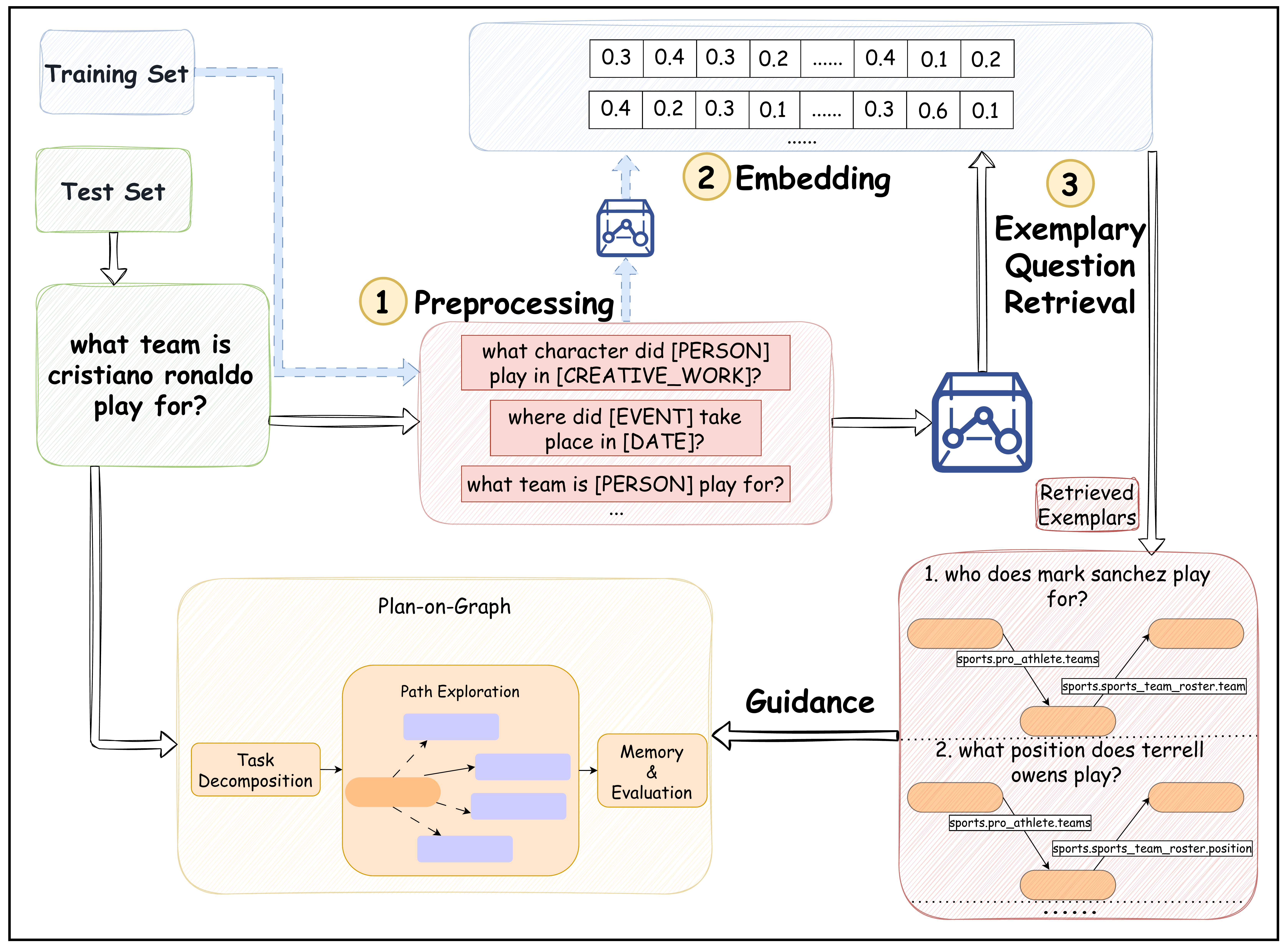}
\caption{Overview of the EGP framework.}
\label{EGP}
\end{figure}

\subsection{EGP framework}

As illustrated in Fig. \ref{EGP}, EGP consists of three key components: question preprocessing, text embedding generation, and exemplary question retrieval. EGP first preprocesses the question texts from the training set, primarily through entity templating. Entity templating aims to mitigate semantic variations across questions caused by differing specific entities. By replacing concrete entities with their category labels, this process promotes more consistent and structurally-focused embeddings for questions that are semantically similar in their underlying intent. Subsequently, a text embedding model(bge-large-en-v1.5 in this paper) is employed to batch-generate embeddings for the preprocessed question texts, forming a training set question embedding matrix. An index for this embedding matrix is then built using FAISS to accelerate retrieval efficiency. When the QA system comes to a new question, it first subjects the question text to the same preprocessing steps as described above. The same text embedding model is then used to obtain a vector representation of the question, which is subsequently used to retrieve the nearest neighbor questions from the training set via the constructed FAISS index. Based on the retrieved question IDs, their corresponding correct reasoning paths are obtained. It is particularly noteworthy that the EGP framework does not solely select the top few exemplary questions based on embedding similarity; we also consider a similarity threshold to enhance the referential quality of the exemplars, as well as the diversity of reasoning paths among the exemplars to avoid providing redundant reference paths and to improve generalization. These retrieved exemplary questions and their reasoning paths play a crucial role in guiding the LLM's interactive exploration on the KG. We apply our method on top of PoG, and the specific details are as follows.

\subsection{Task Decomposition with EGP}

\begin{figure}[tbp]
\centering
\includegraphics[width=0.7\textwidth]{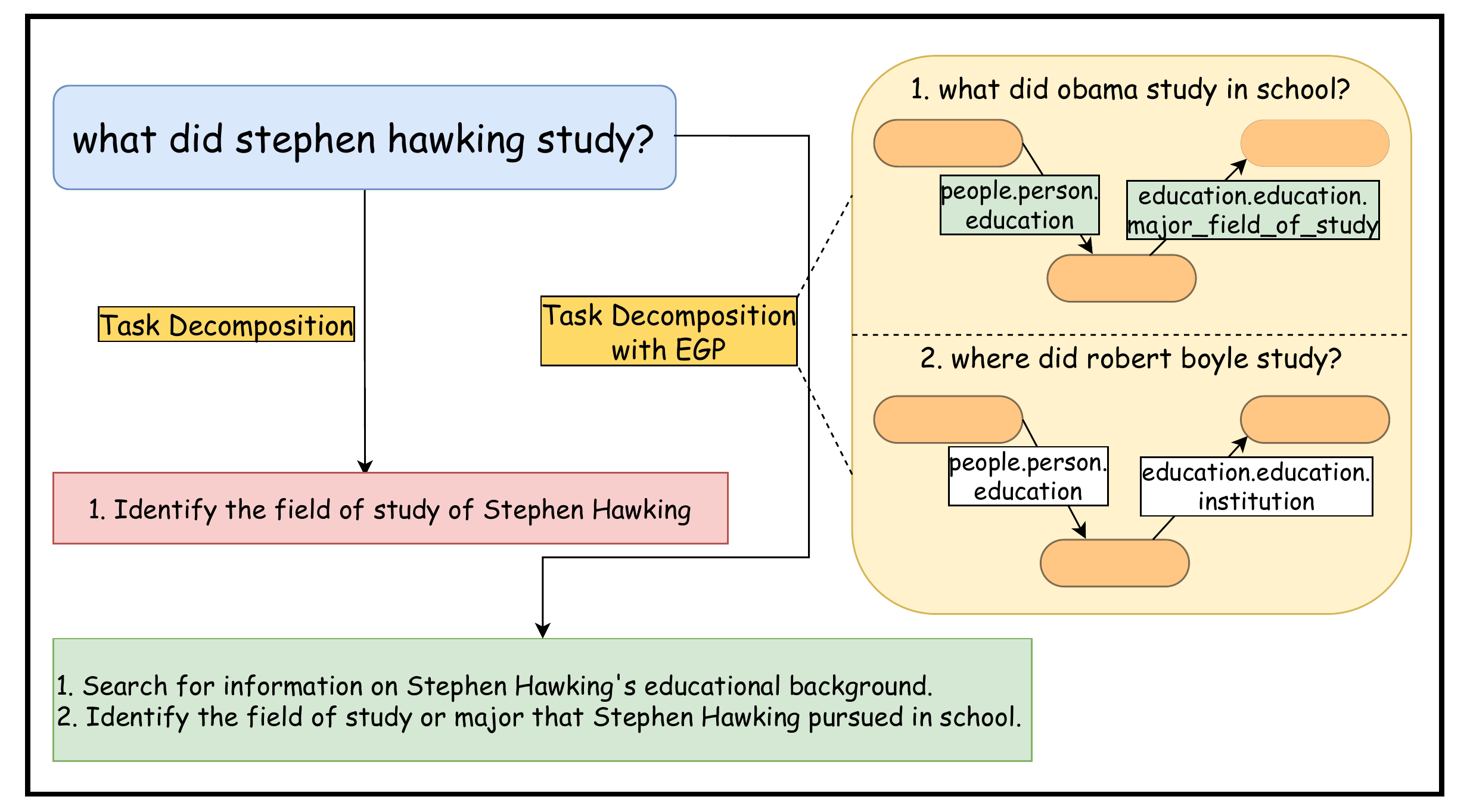}
\caption{An example of task decomposition with EGP.}
\label{TD}
\end{figure}

Task decomposition aims to leverage the LLMs to convert an original question into more executable sub-objectives. The sub-objectives obtained after task decomposition will be used to guide the LLM in subsequent steps. In PoG, task decomposition is performed by directly prompting the LLM with a one-shot example. Such decomposition results often struggle to align with the structural hierarchy within the KG, leading to insufficient guidance effectiveness. With the assistance of EGP, we inject exemplary questions and their correct reasoning paths into the prompt. This enables the LLM to identify relations within the KG that are highly valuable for solving the current question and to perform task decomposition based on these relations. Consequently, the resulting sub-objectives are better aligned with the granularity and semantics of the relation paths used to solve problems in the KG. An example is given in Fig. \ref{TD}. We instruct the LLM to output sub-objectives in a list format, where there is a certain logical connection (usually sequential) between the sub-objectives. Formally, the list of sub-objectives can be represented as $O=\{o_1, o_2, o_3, \ldots\}$.

\subsection{Path Exploration with EGP}

Path exploration involves the LLM continuously and interactively accessing KG information until a reasoning path is discovered to answer the question. The starting point for path exploration is the topic entities in the question. Following previous studies\cite{sun2024think, chen2024pog}, topic entities and their IDs in the KG are assumed to be pre-annotated in the dataset and can be used directly. Formally, the initial set of entities for reasoning paths is initialized as $E^0 = T_q = \{ e_1^0, e_2^0, \ldots, e_{N_0}^0 \}$, where $N_0$ is the number of topic entities in the question. In each iteration, the system sequentially performs a round of relation and entity exploration, thereby continuously delving deeper into the KG structure to find the correct reasoning paths and answers.

Taking the $D$-th iteration as an example, before the iteration begins, each explored reasoning path $p_n \in P$ consists of $D_{p_n}$ ($D_{p_n} \leq D -1$) reasoning steps, i.e., $p_n=e_0 \xrightarrow{r_1} e_1 \xrightarrow{r_2} \cdots \xrightarrow{r_{D_{p_n}}} e_{D_{p_n}}$. It should be noted that the length of each reasoning path in $P$ may vary, because in each iteration, the LLM will only select the reasoning paths most relevant to the question to continue exploring. Before the $D$-th iteration, the set of entities and relations to be explored are denoted as $E^{D-1} = \{ e_1^{D-1}, e_2^{D-1}, \ldots, e_{N_{D-1}}^{D-1} \}$ and $R^{D-1} = \{ r_1^{D-1}, r_2^{D-1}, \ldots, r_{N_{D-1}}^{D-1} \}$, respectively, where $N_{D-1}$ is the length of $E^{D-1}$ and $R^{D-1}$. In each iteration, the LLM will determine which paths in the reasoning path set $P$ need to be extended, along with their corresponding next-hop relations and entities, based on the question $q$, the sub-objectives list $O$, the sets of entities and relations to be explored $E^{D-1}$ and $R^{D-1}$, and other relevant information. In the actual practice of path exploration, the LLM first identifies the neighboring relations most relevant to the current set of entities to be explored. Then it selects the most relevant entities based on these relations, i.e., a two-step process of relation exploration and entity exploration. The EGP framework primarily plays a role in the relation exploration phase.

\begin{figure}[tbp]
\centering
\includegraphics[width=0.8\textwidth]{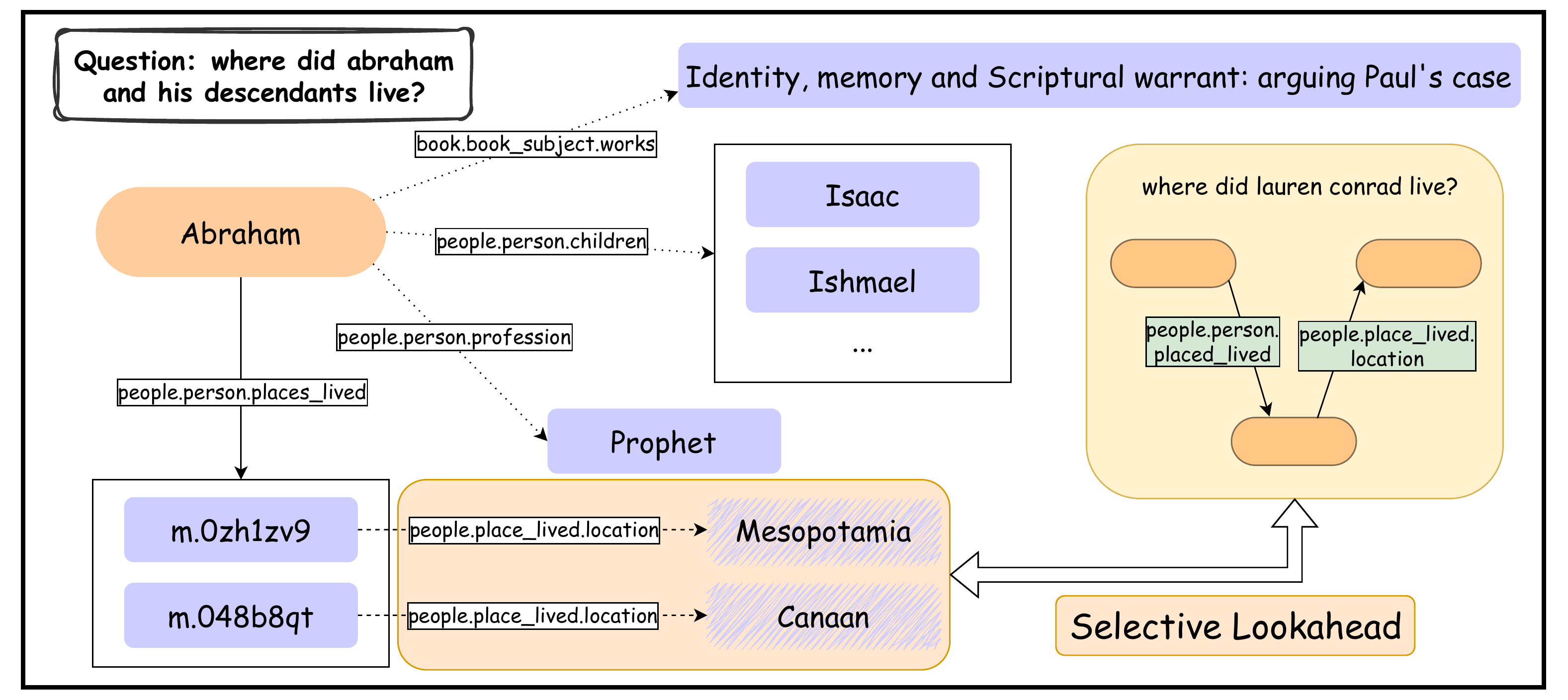}
\caption{An example of path exploration with EGP.}
\label{RE}
\end{figure}

\textbf{Relation Exploration.} Relation exploration aims to prune from the relations connected to entities in $E^{D-1}$ (including both head and tail relations) to obtain a subset of relations most relevant to question $q$ and the sub-objectives list $O$. Specifically, predefined SPARQL queries are first used to retrieve all relations $R_{cand}^{D}$ connected to the entities in $E^{D-1}$. Subsequently, the LLM is prompted with information such as the question $q$, the set of entities to be explored $E^{D-1}$, and the list of candidate relations $R_{cand}^D$, to obtain the relations $R_{sel}^D$ from $R_{cand}^D$ that are most likely to solve the problem. Notably, since the EGP framework has already retrieved several exemplary questions from the training set that are highly similar to the current question, we add these exemplary questions and their corresponding correct reasoning paths as guiding information to specific positions in the prompt for the LLM's reference. To ensure that the exemplary questions and their reasoning paths are fully utilized, we extract all relations $R_{guide}$ that appear in these reasoning paths. The final set of relevant relations obtained from relation exploration is $R^D=R_{sel}^D\cup(R_{cand}^D \cap R_{guide})$. This $R^D$ is then used to extend the current reasoning paths $p_n \in P$, for example, $p_n=e_0 \xrightarrow{r_1} e_1 \xrightarrow{r_2} \cdots \xrightarrow{r_{D_{p_n}}} e_{D_{p_n}} \xrightarrow{r^D_n} (?)$. Furthermore, we incorporate a Smart Lookahead mechanism during the first iteration: if $R_{cand}^1 \cap R_{guide}=R_{forward}\ne \varnothing$, we attempt to explore whether a reasoning path $p_t=e^0_i \xrightarrow{r_t} m \xrightarrow{r_{g_1}} \ldots \xrightarrow{r_{g_n}} e_t$ exists, where $e_i^0 \in E^0$, $r_t \in R_{forward}$, and the relation path $z_t=\{r_t, r_{g_1}, \ldots, r_{g_n}\}$ matches a reasoning path from an exemplary question. Subsequently, the LLM determines if the reasoning path $p_t$ is sufficient to answer the question. If the LLM deems it a correct and complete reasoning path relevant to the question, the answer is given based on $p_t$.

\textbf{Entity Exploration.} Like relation exploration, entity exploration aims to retrieve the entities most relevant to question $q$ based on $E^{D-1}$ and $R^D$. After completing relation exploration, queries such as $(e^{D-1}_n, r^D_n,?)$ or $(?, r^D_n,e^{D-1}_n)$ are executed to obtain candidate entity sets $E^D_{cand,n}$, where $e^{D-1}_n$ and $r^D_n$ are the tail entity and tail relation in $p_n$, respectively. If a candidate entity set $E^D_{cand,n}$ is too large, a certain number of entities will be filtered based on their semantic similarity to question $q$ with a small pre-trained DistilBERT\cite{sanh2019distilbert}. Subsequently, all candidate entities are used to construct $E_{cand}^D$, which is then used to further extend $p_n$. At this point, $p_n=e_0 \xrightarrow{r_1} e_1 \xrightarrow{r_2} \cdots \xrightarrow{r_{D_{p_n}}} e_{D_{p_n}} \xrightarrow{r^D_n} e^D_n$, where $e^D_n \in E^D_{cand}$. Then, based on question $q$ and the set of tail triples $(e^{D-1}_n, r^D_n, e^D_n)$ in $p_n$, the LLM is prompted to select the reasoning paths most relevant to question $q$.

\subsection{Memory \& Evaluation}

Following prior study\cite{chen2024pog}, after each iteration of path exploration, we summarize the acquired information and update it into memory for historical retrieval and reflection. The primary information recorded includes: the explored subgraph, reasoning paths, and the current known status for each sub-objective. 1) \textbf{Subgraph}: This includes all KG triples explored so far, facilitating the identification of entities to backtrack to during subsequent reflection. 2) \textbf{Reasoning Paths}: Beyond the currently explored entities and relations, preceding paths leading to them are preserved. This helps the LLM better understand inter-entity relations and correct erroneous paths during reflection. It also enables a clear and accurate presentation of the reasoning process when an answer is found. 3) \textbf{Sub-Objective Status}: The LLM updates the current status of each sub-objective based on the latest path exploration outcomes and existing memory, aiding the reflection phase.

Upon memory update, the LLM determines if the currently acquired information is sufficient to infer an answer. If the LLM deems the information sufficient, it will provide an answer based on the existing reasoning paths, sub-objective statuses, and its knowledge. If the information is insufficient, the system enters a reflection state to assess if the current reasoning paths are erroneous and require correction. Specifically, the LLM reflects on whether to correct the current exploration direction based on the question $q$, sub-objectives statuses, reasoning paths $P$, and the set of tail entities $E^D$ planned for the next iteration. It also provides a reason for its decision. If the LLM finds it necessary to incorporate previously explored entities (other than those in $E^D$) into the exploration, a self-correction of the current reasoning paths is needed. The LLM will then decide which entities $E_{add}^D$ to backtrack to from the historical candidate entity set $E_{\text{cand}} = E^{1}_{\text{cand}} \cup E^{2}_{\text{cand}} \cup \ldots \cup E^{D}_{\text{cand}}$, and adds them to $E^D$ for re-exploration in the next iteration, i.e., $E^{D} = E^{D} \cup E_{\text{add}}^{D}$.

\section{Experiments}

In this section, we evaluate the effectiveness of the EGP framework on widely used datasets. We answer the following research questions through experiments to demonstrate the efficacy of our method:

\begin{itemize}
    \item \textbf{RQ1:} Does PoG's performance improve after applying the EGP framework?
    \item \textbf{RQ2:} Does the guidance from exemplary questions help improve the accuracy of relation pruning?
    \item \textbf{RQ3:} Does the introduction of the Smart Lookahead mechanism enhance system efficiency?
\end{itemize}

\subsection{Experimental Setups}

\textbf{Dataset and Evaluation.} We employ two representative KGQA datasets for our experiments: WebQSP\cite{yih2016value} and CWQ\cite{talmor2018web}, which contain questions with up to 4 hops. Both datasets are constructed based on the FreeBase knowledge base\cite{bollacker2008freebase}, which comprises approximately 88 million entities, 20,000 relations, and 126 million triples. It is important to note that since the CWQ dataset does not contain explicit reasoning path information, we used an LLM to generate reasoning paths based on the correct SPARQL query for each question. We adopted exact match accuracy (Hits@1) as the evaluation metric for ease of comparison.

\textbf{Baselines.} We select several representative SOTA methods as baselines, primarily focusing on comparing against the PoG method without the EGP framework. The baseline methods can be categorized into two groups: (1) LLM-only methods, including standard prompting (IO prompt)\cite{brown2020language}, Chain-of-Thought (CoT)\cite{wei2022chain}, and Self-Consistency (SC)\cite{wang2023self}; (2) KG-augmented LLM methods, encompassing both fine-tuned and prompting approaches. For fine-tuned methods, we primarily compare against UniKGQA\cite{jiang2023unikgqa}, TIARA\cite{shu2022tiara}, RE-KBQA\cite{cao2023pay}, DeCAF\cite{yu2023decaf}, and RoG\cite{luo2024reasoning}. For prompting methods, we compare against KD-CoT\cite{wang2023knowledge}, KB-BINDER\cite{li2023few}, StructGPT\cite{jiang2023structgpt}, InteractiveKBQA\cite{xiong2024interactive} and PoG\cite{chen2024pog}.

\textbf{Foundational LLM.} We conduct our experiments using two LLMs: GPT-3.5 (specifically, gpt-3.5-turbo) and gemini-2.5-flash-preview-0417. The use of GPT-3.5 is to maintain consistency with previous experiments for comparison purposes. The choice of gemini-2.5-flash-preview-0417 is based on a balance between invocation cost and model performance, selecting a powerful LLM to investigate the current upper bound of our method's capabilities.

\subsection{RQ1: KGQA Performance}

\table
\small
\centering

\caption{Performance comparison on WebQSP and CWQ.}
\label{Tab:main}
\setlength{\tabcolsep}{0.5mm}{\begin{tabular}{lcc}
\toprule
\textbf{Method}  & \textbf{WebQSP} & \textbf{CWQ}  \\ \midrule
\multicolumn{3}{c}{\textit{LLM-Only}} \\ \midrule
IO Prompt~\cite{brown2020language}&63.3&37.6\\
CoT~\cite{wei2022chain}&62.2&38.8\\
SC~\cite{wang2023self}&61.1&45.4\\
\midrule
\multicolumn{3}{c}{\textit{Fine-Tuned KG-Augmented LLM}} \\ \midrule
UniKGQA~\cite{jiang2023unikgqa} & 79.1 & 51.2\\
TIARA~\cite{shu2022tiara} & 75.2&-\\
RE-KBQA~\cite{cao2023pay} & 74.6 & 50.3\\
DeCAF~\cite{yu2023decaf} & 82.1&70.4\\
RoG~\cite{luo2024reasoning} & 85.7 &62.6\\
 \midrule
\multicolumn{3}{c}{\textit{Prompting KG-Augmented LLM w/GPT-3.5 or others}} \\ \midrule

KD-CoT~\cite{wang2023knowledge}& 73.7 &50.5 \\
KB-BINDER~\cite{li2023few}&74.4&-\\
StructGPT~\cite{jiang2023structgpt} & 72.6 & 54.3\\
PoG~\cite{chen2024pog}(replicated)& 79.7 & 60.2\\
\textbf{PoG-EGP}& \textbf{83.6} &  \textbf{63.8} \\
\midrule
\multicolumn{3}{c}{\textit{Prompting KG-Augmented LLM w/GPT-4}}\\
\midrule
InteractiveKBQA~\cite{xiong2024interactive} & 72.5 &59.2\\
PoG~\cite{chen2024pog}   & 87.3 & 75.0\\
\midrule
\multicolumn{3}{c}{\textit{Prompting KG-Augmented LLM w/Gemini-2.5-Flash-Preview-0417}}\\
\midrule
\textbf{PoG-EGP}   & \textbf{88.6} &\textbf{75.4}
\\ \bottomrule
\end{tabular}}
\endtable

\begin{figure}[tbp]
\centering
\includegraphics[width=0.8\textwidth]{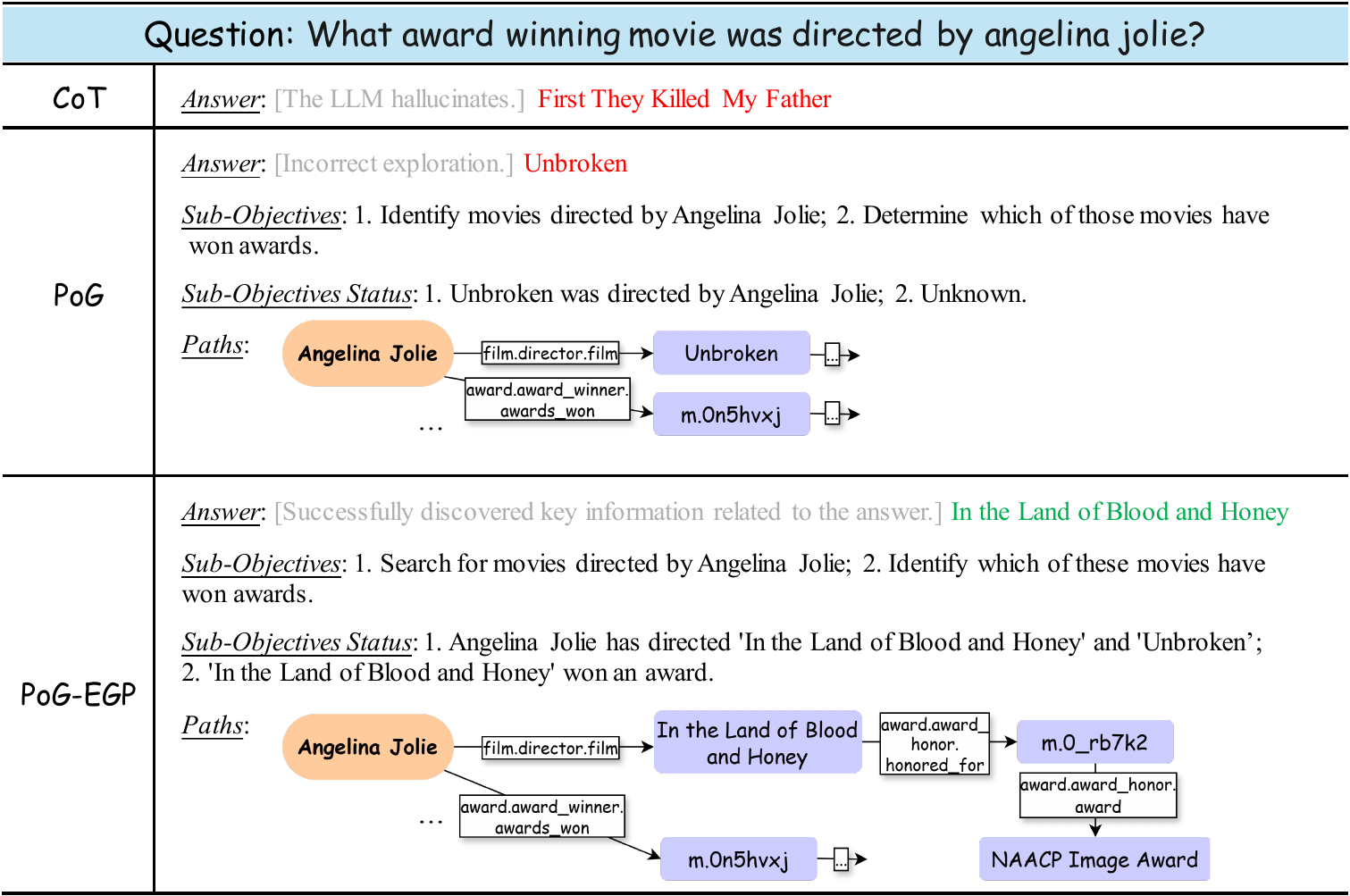}
\caption{Case Study.}
\label{CS}
\end{figure}

\table
\small
\centering

\caption{Ablation study of \textit{EGP}.}
\label{Tab:ab}
\setlength{\tabcolsep}{0.5mm}{\begin{tabular}{lcc}
\toprule
\textbf{Method}  & \textbf{WebQSP} & \textbf{CWQ}\\ \midrule
\textbf{PoG-EGP}   & \textbf{83.6} &\textbf{63.8} \\ 
w random exemplars &  80.4 & 61.0\\ 
w/o guidance in task decomposition & 83.1 & 62.4\\ 
w/o guidance in path exploration & 81.7 & 61.5\\
\bottomrule
\end{tabular}}
\endtable

\textbf{Main Results.} Table \ref{Tab:main} shows PoG-EGP outperforms all baselines across both datasets, with the strongest results achieved using gemini-2.5-flash-preview-0417 as the base LLM. Compared to PoG, our method improves performance by 3.9\% and 3.6\% on WebQSP and CWQ respectively with GPT-3.5. PoG-EGP surpasses the best baseline by 1.3\% on WebQSP and 0.4\% on CWQ. Notably, our training-free approach with offline index construction introduces minimal overhead while achieving comparable or superior results to fine-tuned LLM methods. A case study, presented in Fig. \ref{CS}, further illustrates the enhanced planning and reasoning capabilities of PoG-EGP.

\textbf{Ablation Study.} We conducted ablations on WebQSP and CWQ using GPT-3.5 as shown in Table \ref{Tab:ab}. Three variants were tested: (1) \textit{w/ random exemplars}: random exemplars instead of similarity-based retrieval, (2) \textit{w/o guidance in task decomposition}: removing guidance from task decomposition, and (3) \textit{w/o guidance in path exploration}: removing guidance from path exploration. All variants showed performance decreases, with random exemplars causing the most significant drop. Guidance proved more critical in the path exploration phase, suggesting that guidance from the exemplars helps LLMs better understand the logical structure of KG and the question.

\subsection{RQ2: Relation Pruning with EGP}

\begin{figure}[htbp]
\centering
\includegraphics[width=0.5\textwidth]{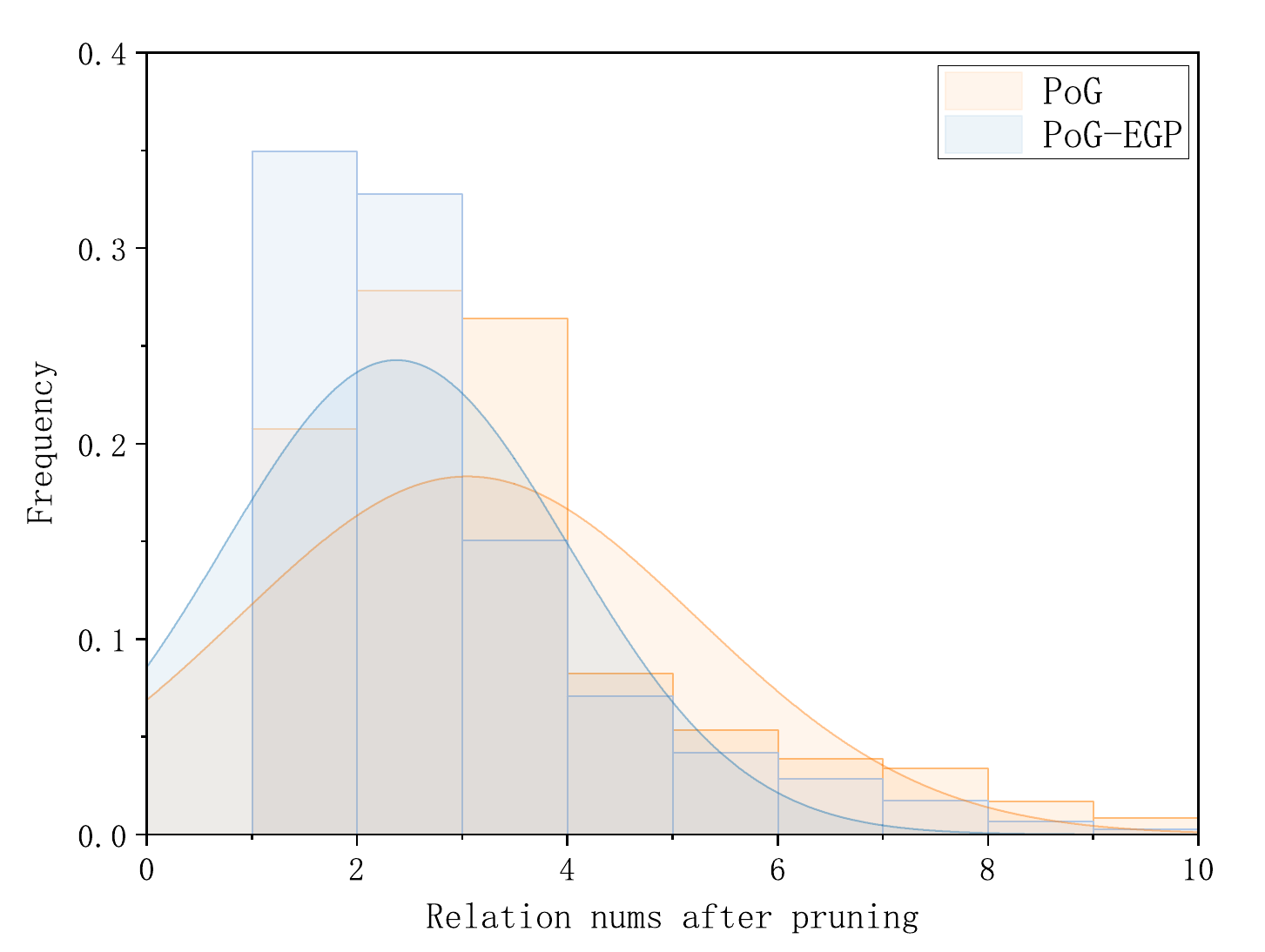}
\caption{Relation nums comparison.}
\label{RQ2}
\end{figure}

This section investigates whether the EGP framework effectively guides the LLM's relation pruning. To this end, during the relation pruning process, we counted the number of relations remaining after pruning, both with and without the guidance of exemplary questions, and then performed a frequency distribution analysis with curve fitting. The final experimental results are illustrated in Fig. \ref{RQ2}. We can observe that the two probability distribution curves show a relatively clear distinction. This indicates that the number of remaining relations is generally higher when LLM performs relation pruning based on a simple prompt without guidance from the exemplars. After introducing exemplary question guidance, this number decreases. Furthermore, based on the system performance experiments in RQ1, we believe that, overall, after guiding the LLM's relation pruning with exemplary questions, the LLM's perspective becomes more focused and accurate during KG exploration. This plays an important role in avoiding exploration into irrelevant branches.

\subsection{RQ3: Effectiveness of Smart Lookahead Mechanism}

This section discusses the effectiveness of the introduced Smart Lookahead mechanism. In practice, the Smart Lookahead mechanism introduces a small number of additional LLM calls, so we aim to observe whether the Smart Lookahead mechanism can terminate KG exploration prematurely in most cases, thereby reducing subsequent overhead. To this end, we counted the number of times the Smart Lookahead mechanism was triggered and the number of times questions were correctly answered prematurely based on the Smart Lookahead mechanism. We found that in the WebQSP dataset, the Smart Lookahead mechanism was triggered 1407 times, accounting for approximately 85.8\% of the total test set. Of these triggered instances, 864 questions (61.4\%) were correctly answered prematurely due to the mechanism. In the CWQ dataset, the corresponding results were 89.2\% and 26.0\%, respectively. This demonstrates the effectiveness of the Smart Lookahead mechanism.

\subsection{Conclusion}

In this paper, we propose a novel exemplar-guided enhancement framework, EGP, which introduces knowledge from the training set regarding the current question and KG structure to the LLM, thereby improving the planning capability of LLMs in interactive KG reasoning methods. EGP preprocesses question texts through entity templating, then uses a text embedding model and the FAISS library to build an index for retrieving exemplary questions from the training set. We applied EGP to the PoG base method, termed PoG-EGP. Specifically, PoG-EGP incorporates auxiliary information from EGP into the task decomposition and relation exploration phases, and also designs a Smart Lookahead mechanism to enhance system efficiency. On two widely recognized datasets, WebQSP and CWQ, the PoG-EGP method achieved optimal performance compared to 13 baseline models. We also conducted extensive experiments to demonstrate the effectiveness of the EGP framework.

\textbf{Acknowledgements.} This work was supported by the National Key Research and Development Program of China (Grant No.2023YFC3306104).

\bibliographystyle{splncs04}
\bibliography{ref.bib}
\end{document}